\documentclass[journal]{IEEEtran}
\usepackage{graphicx} 
\usepackage{hyperref}
\usepackage{booktabs}
\usepackage{makecell}
\usepackage{soul}
\usepackage{xcolor}
\usepackage{makecell}
\usepackage{siunitx}
\usepackage{cleveref}
\usepackage{caption}
\usepackage{subcaption}

\usepackage{amssymb}

\title{Code-mixed Sentiment and Hate-speech Prediction}
\author{Anjali Yadav, Tanya Garg, Matej Klemen, Matej Ulčar, \\ Basant Agarwal, Marko Robnik-Šikonja}
\date{July 2023}

\begin{document}

\maketitle

\begin{abstract}


Code-mixed discourse combines multiple languages in a single text. It is commonly used in informal discourse in countries with several official languages, but also in many other countries in combination with English or neighboring languages. As recently large language models have dominated most natural language processing tasks, we investigated their performance in code-mixed settings for relevant tasks. We first created four new bilingual pre-trained masked language models for English-Hindi and English-Slovene languages, specifically aimed to support informal language. Then we performed an evaluation of monolingual, bilingual, few-lingual, and massively multilingual models on several languages, using two tasks that frequently contain code-mixed text, in particular, sentiment analysis and offensive language detection in social media texts. The results show that the most successful classifiers are fine-tuned bilingual models and multilingual models, specialized for social media texts, followed by non-specialized massively multilingual and monolingual models, while huge generative models are not competitive. For our affective problems, the models mostly perform slightly better on code-mixed data compared to non-code-mixed data.

\end{abstract}

\section{Introduction} 

Code-mixing, the practice of combining multiple languages or varieties in a single discourse, is common in today's world. This phenomenon is observed in bilingual and multilingual communities, and influenced by informal settings, cultural identity, lack of vocabulary, media and pop culture, globalization, and the digital era \cite{thara2018code}. In bilingual households, people often switch between languages based on context or personal preference. In informal settings, code-mixing is more common when speaking with friends or expressing ideas or emotions in a specific language \cite{thara2018code}. It can also be a way to express cultural identity or adapt to linguistic diversity in multicultural environments. The digital era has introduced new avenues for communication, making code-mixing more prevalent \cite{androutsopoulos2011codeswitching}.

Analyses of code-mixed data enable understanding of realistic interactions in real-world communication, which is especially important for understanding sentiment and emotions, expressed by people in multilingual environments. Successful analysis of code-mixed data also provides effective multilingual insights and enhances decision-making in various domains like marketing \cite{kathpalia2015use}, customer support \cite{SCHAU200765}, and policy-making \cite{sravani2021political}. 

Recently, pre-trained large language models (LLMs) have dominated the landscape of Natural Language Processing (NLP). LLMs appear in various sizes, from 100 million to several hundred billion parameters, and cover different numbers of languages, from monolingual to massively multilingual, supporting a few hundred languages. In the context of code-mixing, two types of LLMs are interesting: those covering all the languages that simultaneously appear in a code-mixed text, and those pre-trained on informal texts, which are the most frequently used in code-mixed scenarios. While several works applied LLMs to tasks with code-mixed languages (see Section \ref{sec:related} for an overview), the existing analyses were limited to individual problems and pairs of code-mixed languages. We fill this gap and conduct an analysis, covering multiple types of LLMs on five languages (French, Hindi, Russian, Slovene, and Tamil).  We tackle two affective computing tasks where code-mixing is especially prominent: sentiment analysis and hate speech detection. 

Analyses of sentiment and hate speech detection in a code-mixed setting are particularly relevant for affective computing due to increased cultural and linguistic diversity in a globalized world. In this context, modern affective computing systems must handle and interpret inputs from diverse linguistic backgrounds, including those where code-mixing is a natural part of communication. Emotions can be expressed differently across languages, and the nuances might change significantly in code-mixed environments. Models trained on monolingual data often fail to capture these subtleties. Adapting them to handle code-mixed text might improve their performance, robustness, user experience, and inclusiveness, especially in informal settings. Our study builds upon the growing interest in sentiment analysis of code-mixed text. A recent study introduced robust transformer-based algorithms to enhance sentiment prediction in code-mixed text, specifically focusing on English and Roman Urdu combinations.\cite{hashmi2024augmenting} The study employed state-of-the-art transformer-based models like Electra\cite{clark2020electra}, code-mixed BERT (cm-BERT), and Multilingual Bidirectional and Auto-Regressive Transformers (mBART).\cite{dominic2023multilingual}

The aim of our work is to investigate the abilities of different large language models, which nowadays represent the essential infrastructure for language analysis, in the area of code-mixed sentiment and hate speech. 

Our main contributions are as follows:
\begin{itemize}
    \item We created four new masked bilingual large language models, focused on informal language.
    \item We conducted an in-depth analysis of large language models on code-mixed language using two types of representative classification problems in five languages, where such problems are relevant and adequate large language models exist. The results show an advantage of bilingual models and models specialized for social media texts over general massively multilingual and monolingual models.
    \item We investigated the detection of code-mixed language and observed weaknesses in existing commonly used language detectors.
\end{itemize}

The remainder of our paper is structured as follows.
In \ ref{sec:related}, we review the related literature on code-mixing patterns, language models, and code-mixed evaluation. 
In \Cref{sec:datasets}, we describe the background and essential statistics of datasets we use in our code-mixed evaluation.
In \Cref{sec:Pre}, we present training of language models used in our study, followed by the description and analysis of the experimental results.
We conclude in \Cref{sec:conclusion} with an overview of the findings, a discussion on the limitations of the study, and possible future directions for code-mixed natural language processing.

\section{Related work}
\label{sec:related}
In this section, we overview the related literature on code-mixing. Initially, we present an overview of multilingual LLMs that are suitable for modeling code-mixed language in \Cref{sec:code-mixed-llms}. We describe the existing resources used to evaluate code-mixed NLP in \Cref{sec:benchmarks-metrics-for-code-mixed}, and studies on code-mixed evaluation of LLMs in \Cref{sec:code-mixed-evaluations}. Finally, in \Cref{sec:affective-computing-code-mixed}, we present the existing works connecting affective computing tasks and code-mixing.


\subsection{Multilingual LLMs}
\label{sec:code-mixed-llms}
Initially, large language models (LLMs) were primarily trained on well-resourced languages like English and Chinese due to the availability of ample resources. However, there was a limitation in understanding text containing multiple languages simultaneously. Many widely spoken languages lacked sufficient resources for model training. Multilingual LLMs such as the multilingual BERT \cite{bert} and XLM-RoBERTa \cite{xlm-roberta} were developed to address this issue. These models have been trained on balanced datasets comprising about 100 languages, aiming to provide better support for multilingual text understanding and processing.

Similarly, several few-lingual LLMs have been developed, each trained on a limited number of languages. Examples include CroSloEngual BERT (supporting Croatian, Slovene, and English), FinEst BERT (trained on Finnish, Estonian, and English) \cite{csebert},  and MuRIL \cite{khanuja2021muril} (supporting 17 Indian languages)

For understanding code mixing, models need to comprehend how different languages are interchangeably occurring within a single sentence, typically in informal discourse and often used in an affective context. While multilingual models are pre-trained on general language data (e.g., Wikipedia dumps), code-mixing presents a specialized challenge. An alternative and direct way of improving the code-mixed performance is by pre-training models directly on code-mixed texts. For example, Nayak and Joshi release multiple transformer-based models trained on a carefully curated Hindi-English corpus \cite{nayak-joshi-2022-l3cube}. Such models are aimed at handling the specifics of code-mixing but are not available for many language pairs.


In this work, we extend the support for code-mixed language processing by introducing four new specialized LLMs: two for the previously unsupported Slovene-English language pair, and two for Hindi-English. Additionally, we perform a comprehensive code-mixed evaluation using a selection of LLMs, including few-lingual models previously untested in such scenarios. 

\subsection{Evaluation in code-mixed settings}
\label{sec:benchmarks-metrics-for-code-mixed}

Code-mixed LLMs are typically evaluated through extrinsic evaluation, i.e. by evaluating the system on a downstream task that involves code-mixing. The tasks depend on the end goal and are either generative or discriminative. Examples of generative tasks include code-switched text translation and controlled code-switched text generation \cite{phinc-hinglish-code-mixed-mt, controlled-cm-generation}, while examples of discriminative tasks include offensive language identification, sentiment analysis, and natural language inference \cite{code-mixed-nli, Chakravarthi2022}. To provide a more general evaluation, authors have also released benchmarks spanning multiple tasks \cite{gluecos}. 
Likely due to its informal setting, the code-mixed datasets are commonly sourced from social networks (e.g., Twitter) or other online platforms with comment sections. 
Despite the ubiquitous nature of code-mixing, the pool of languages used in code-mixing evaluations is relatively small and typically involves a language code-mixed with English, for example, one of the Indo-Aryan or Dravidian languages \cite{indoaryan-dravidian-qa}; alternatively, a higher-resourced language such as Spanish \cite{spanish-msa-ea-ner} can also be code mixed with English. However, code-mixing does not necessarily occur paired with English, and some resources have been collected for such settings: for example, the Turkish-German dependency parsing dataset \cite{turkish-german-ud}, or the modern standard Arabic and Egyptian dialectal Arabic dataset \cite{msa-ea-dataset}.

In our work, we focus on the evaluation of LLMs on two affective tasks: offensive language identification and sentiment analysis. In contrast to existing work, we consider a larger pool of languages (five language pairs), including languages for which code-mixed research is scarce.

\subsection{Comprehensive analyses of code-mixing}
\label{sec:code-mixed-evaluations}

Most works mentioned in \Cref{sec:benchmarks-metrics-for-code-mixed} have focused on evaluating and optimizing LLMs for a specific code-mixed task or a small pool of tasks in one language. 
Several works have instead focused on a larger-scale evaluations, which we focus on in our work as well.

Winata et al. \cite{winata2021multilingual} study the effectiveness of multilingual LLMs on code-mixed named entity recognition and part of speech tagging across three language pairs using three criteria: inference speed, performance, and number of parameters. Zhang et al. \cite{zhang-etal-2023-multilingual} perform a similar analysis, comparing the performance of few-hundred million parameters fine-tuned massively multilingual LLMs to the performance of multi-billion parameters models in zero-shot and few-shot settings. They find that while large models are relatively successful, fine-tuned smaller models achieve the best results. 
Santy et al. \cite{santy2021bertologicomix} study the effect of using different types of synthetic data to fine-tune multilingual LLMs for six tasks across one or two language pairs. They find that including any type of sythetic code-mixed data in the tuning process improves the responsivity of attention heads to code-mixed data, indicating the suboptimal support for code-mixing in multilingual models. 
Tan and Joty \cite{tan-joty-2021-adversarial} and Birshert and Artemova \cite{birshert2022adversarial} study the capability in an adversarial setting, and show that synthetically constructed code-mixed examples cause a significant drop in the accuracy of multilingual LLMs. 

In our work, we aim to continue the line of comprehensive analyses on a pool of five languages, including multiple that are not commonly studied in the existing literature. We focus our analysis on two affective evaluation tasks and evaluate a large pool of monolingual, few-lingual, and massively multilingual LLMs.

\subsection{Affective Computing in Code-Mixed Language Modeling}
\label{sec:affective-computing-code-mixed}

Affective computing, incorporating emotion and sentiment comprehension in human language, is an important area of NLP.  aiming to decipher and understand emotional nuances\cite{picard2000affective}. Yet, when it comes to code-mixed content, the landscape becomes more intricate.

A primary obstacle to better understanding affective hints arises from the fact that emotions can be deeply intertwined with the choice of language for particular words or phrases. For example, embedding a term of endearment in Hindi within an English sentence can completely shift the sentiment, transforming a neutral statement into an affectionate one. For example, consider the English neutral sentence "I received a nice gift today" and the code-mixed sentence "I received a nice gift today, meri pyaari maa". In the original sentence, "nice" conveys a positive but neutral sentiment about the gift. However, in the code-mixed sentence, adding the Hindi phrase "meri pyaari maa" completely transforms the tone.  "meri pyaari maa" expresses affection and suggests the gift has a deeper meaning because it came from the mother. Another example with respect to hate speech is "This politician is a complete bekaar" where bekaar is Hindi for 'useless'. The English sentence criticizes the politician, but "bekaar" adds a stronger layer of insult specific to Hindi. Hate speech detection models trained only on English might miss the hateful connotation because they wouldn't understand the severity of the Hindi word.

Several works have ventured into understanding sentiment analysis and emotion discernment in code-mixed contexts. Balahur et al. \cite{balahur2011emotinet} introduce a model that leverages cultural context and societal norms to enhance emotion detection precision. Additionally, Bedi et al.\cite{bedi2021multi} probe the intricacies of spotting sarcasm in code-mixed exchanges, emphasizing the importance of grasping the dynamics between languages.

\section{Code-mixed Affective Datasets}
\label{sec:datasets}

In this section, we present ten affective code-mixed datasets used to test the newly introduced and existing LLMs. For each language, we select one sentiment analysis dataset and one offensive language detection dataset. Their summary is presented in Table \ref{tab:dataset-statistics}.
We start by describing the datasets grouped by the primary language in Sections \ref{sec:dataset-French} - \ref{sec:dataset-Russian}. Then, we focus on the label distribution (\Cref{sec:label-distribution}) and the degree of code-mixed content present in the datasets (\Cref{sec:code-mixing-proportion}).

\begin{table*}[htb]
\caption{Summary of the used datasets. The \% of code-mixing is manually estimated on 100 random samples, except for the Hindi and Russian dataset (marked with *), where we used the CodeSwitch and langdetect libraries; see \Cref{sec:code-mixing-proportion} for details. } 
\label{tab:dataset-statistics}
\centering
\begin{tabular}{lcccrr}
\toprule
Dataset & Detected Languages & Task & Genre(s) & \#Inst. & Code-mixing \%\\
\midrule
FrenchBookReviews & French, English & sentiment analysis & book reviews & $9\,658$ & 17.00\\ 

FrenchOLID & French, English & offensive language & tweets & $5\,786$ & 11.00 \\ 

TamilCMSenti & Tamil, English & sentiment analysis & YouTube comments & 36\,681 & 46.00 \\ 

TamilCMOLID & Tamil, English & offensive language & YouTube comments & 41\,760 & 31.00 \\

IIITH & Hindi, English & sentiment analysis & Facebook comments & 3\,879 & *63.88 \\

Hinglish Hate  & Hindi, English & offensive language & tweets & 4\,578 & *86.57\\


Sentiment15\textsubscript{SL} & Slovene, Croatian, English & sentiment analysis & tweets & $87\,392$ & $36.00$ \\ 

IMSyPP\textsubscript{SL} & Slovene, Croatian, English & offensive language & tweets & $47\,538$ & $29.00$ \\ 

Sentiment15\textsubscript{RU} & Russian, Bulgarian & sentiment analysis & tweets & $86\,948$ & *9.15 \\

RussianOLID & Russian, Bulgarian & offensive language & social media comments & $14\,412$ & *3.91 \\
\bottomrule
\end{tabular}
\end{table*}

\subsection{French Datasetss}
\label{sec:dataset-French}

\subsubsection{FrenchBookReviews}
\label{sec:dataset-FrenchSenti}

The French book reviews dataset\footnote{\href{https://www.kaggle.com/datasets/abireltaief/books-reviews}{https://www.kaggle.com/datasets/abireltaief/books-reviews}} contains $9\,658$ reviews by book readers made on the French websites Babelio and Critiques Libres. The sentiment is derived from a five-star rating system used in the review process: reviews with a rating $\leq$ 2.5 are considered negative, reviews with a rating $\geq$ 4.0 are considered positive, and the others are considered neutral. The dataset contains genuine code-mixing examples in French reviews contributed by individuals. 

\subsubsection{FrenchOLID}
\label{sec:dataset-FrenchOLID}

The French offensive language identification dataset \cite{bertweetfr} contains $5\,786$ tweets posted during the COVID-19 pandemic. The authors consider a tweet offensive if the offense is directed towards somebody. Although the dataset is not specifically introduced as code-mixed, we use it as Twitter commonly contains informal language and code-mixing.

\subsection{Tamil datasets}
\label{sec:dataset-Tamil}

The DravidianCodeMix dataset collection \cite{Chakravarthi2022} contains YouTube comments in three Dravidian languages (Tamil, Kannada, and Malayalam) annotated for sentiment analysis and offensive language identification. In our work, we decided to use only one Dravidian language, i.e. Tamil, as our LLMs are covering it. The Tamil dataset contains around $44,000$ examples. The dataset was annotated by between two and five student annotators:
\begin{itemize}
    \item In the sentiment analysis dataset, the data was annotated as positive, neutral, negative, or mixed feelings. We decided to discard the examples annotated as ``mixed feelings'' to maintain a unified three-label sentiment scheme across all our tested languages.
    \item In the offensive language identification dataset, the data was annotated as not offensive, untargeted offensive, or targeted offensive (three options). We consider any type of offensiveness as the positive class, and the rest as the negative class.
\end{itemize}
If the example did not contain the Tamil language, it was labeled as ``not-Tamil''; in our experiments, we discard such examples. 
We refer to the sentiment analysis dataset as TamilCMSenti, and the offensive language identification dataset as TamilCMOLID.

\subsection{Hindi datasets}
\label{sec:dataset-Hindi}

\subsubsection{IIITH}
\label{sec:dataset-HindiSenti}

The IIITH Hindi-English code-mixed sentiment dataset \cite{joshi2016towards} contains $3\,879$ user comments sourced from popular Indian Facebook pages associated with influential figures Salman Khan and Narendra Modi. The examples were annotated by two annotators using a three-level polarity scale (positive, negative, or neutral); only the examples with matching annotations were included in the final dataset.

\subsubsection{Hinglish Hate}
\label{sec:dataset-HindiOLID}

The Hinglish Hate dataset \cite{bohra2018dataset} contains Hindi-English code-mixed social media texts, consisting of $4\,578$ tweets. The authors annotated the dataset with the language at the word level and with the class the tweets belong to (Hate speech or Normal speech)


\subsection{Slovene datasets}
\label{sec:dataset-Slovene}
\subsubsection{Sentiment15\textsubscript{SL}}
\label{sec:dataset-SloveneSenti}

The Sentiment15\textsubscript{SL} \cite{sentiment15} is the Slovene subset of the corpus of over 1.6 million tweets belonging to 15 European languages. The $102\,392$ Slovene tweets were posted between April 2013 and February 2015, and collected using Twitter Search API by constraining the geolocation of the tweet. They were annotated using a standard three-class annotation scheme (positive, neutral, or negative) by seven annotators.

\subsubsection{IMSyPP\textsubscript{SL}}
\label{sec:dataset-SloveneOLID}

IMSyPP\textsubscript{SL} \cite{imsypp} is a Slovene dataset containing tweets posted between December 2017 and August 2020. The tweets were manually annotated twice for hate speech type and target: we consider tweets containing any type of hate speech as positive and others as negative. We preprocessed the data, keeping only tweets where both its annotations agree, in total $47\,538$ examples.

\subsection{Russian datasets}
\label{sec:dataset-Russian}
\subsubsection{Sentiment15\textsubscript{RU}}
\label{sec:dataset-RussianSenti}

The Sentiment15\textsubscript{RU} is the Russian subset of the Sentiment15 corpus of tweets. As the dataset was collected as a whole, its collection is similar to the collection of the Slovene Sentiment15\textsubscript{SL} subset (see \Cref{sec:dataset-SloveneSenti}). In total, the dataset contains $87\,384$ examples.

\subsubsection{RussianOLID}
\label{sec:dataset-RussianOLID}

The Russian language toxic comments dataset (RussianOLID) contains $14\,412$ comments from Russian websites 2ch and Pikabu. The dataset was originally published on Kaggle, with its annotation quality later being independently validated by Smetanin \cite{russianolid}. The texts are annotated for toxicity using a binary annotation scheme (toxic or non-toxic).

\subsection{Label distribution in the code-mixed datasets}
\label{sec:label-distribution}

To provide insight into the used datasets, we quantify their label distribution in \Cref{tab:label-distribution}, separately for sentiment and offensive language datasets.
%
As the numbers show, all datasets are imbalanced to some degree. For offensive language identification, all datasets contain a higher proportion of non-offensive than offensive examples. For sentiment analysis, FrenchBookReviews and TamilCMSenti lean heavily towards positive sentiment, in IIITH the positive sentiment is dominant to a lesser degree, while datasets Sentiment15-SL and Sentiment15-RU show a relatively balanced mix of positive and negative labels with the largest class being neutral. 

\begin{table}[htb]
\caption{Label distributions in the code-mixed datasets.}
\label{tab:label-distribution}

\begin{subtable}{\columnwidth}
\caption{Sentiment analysis datasets.}
\label{tab:label-distribution-sentiment}
\centering
\begin{tabular}{lccc}
\toprule
Dataset name &  positive & neutral & negative \\
\midrule

IIITH & 34.85\% & 50.45\% & 14.70\% \\[5pt]

FrenchBookReviews & 69.06\% & 22.04\% & 8.90\% \\[5pt]

TamilCMSenti & 67.11\% & 18.67\% & 14.20\% \\[5pt]

Sentiment15\textsubscript{SL} & $27.12\%$ & $43.16\%$ & $29.72\%$ \\[5pt]

Sentiment15\textsubscript{RU} & 27.92\% & 40.08\% & 32.00\% \\

\bottomrule
\end{tabular}
\end{subtable}

\vspace{10pt}

\begin{subtable}{\columnwidth}
\caption{Offensive language identification datasets.}
\label{tab:label-distribution-olid}
\centering
\begin{tabular}{lcc}
\toprule
Dataset name & not offensive & offensive \\
\midrule

FrenchOLID & $77.51\%$ & $22.49\%$ \\[5pt]

TamilCMOLID & 75.40\% & 24.60\% \\[5pt]

Hinglish Hate & 63.73\% & 36.27\% \\[5pt]

IMSyPP\textsubscript{SL} & $66.58\%$ & $33.42\%$ \\[5pt]

RussianOLID & $66.51\%$ & $33.49\%$ \\

\bottomrule
\end{tabular}
\end{subtable}

\end{table}

\subsection{Language proportions in code-mixed datasets}
\label{sec:code-mixing-proportion}
An important aspect of code-mixed datasets is the language diversity they contain. 
%
%
%
We used two methods for detecting code-mixing. The first one tackles code-mixing in French, Russian, Tamil, and Slovene tweets using the \emph{langdetect}\footnote{\href{https://pypi.org/project/langdetect/}{https://pypi.org/project/langdetect/}} library. This method first removes non-alphabetic characters from the text and then uses a character n-gram-based naïve Bayes language detector. We obtain a list of detected languages for each text and consider them code-mixed if more than one language is detected above the threshold of 5\%.
We selected this threshold by analyzing the accuracy of the langdetect library on hand-picked samples, 50 actually code-mixed and 50 not, from each of the aforementioned datasets. We observed the peak accuracy for the 4-7\% threshold range. Balancing precision and recall, we chose 5\% to reliably identify code-mixed content while minimizing false positives and false negatives. 


Due to problems with code-switching identification in Hindi datasets, and the availability of a better alternative, we tried the \emph{CodeSwitch} library\footnote{\href{https://pypi.org/project/codeswitch/}{https://pypi.org/project/codeswitch/}} for this language. The library is based on multilingual BERT, which is known to effectively take the context of neighboring words into account. We used its module, configured specifically for Hindi and English (hin-eng), to identify languages within each tweet. The approach identified Hindi, English, Nepali, and other languages.
Upon manual examination of a subset of labeled instances, we observed that Hindi and Nepali were often identified interchangeably due to their linguistic similarities and shared vocabulary. To enhance the accuracy of our language identification process, we refined our function to categorize a tweet as code-mixed only when it contained combinations of Hindi and English, Nepali and English, or a trilingual mix of Hindi, Nepali, and English. Tweets exhibiting a single language were labeled as 'not code-mixed'.



\begin{table}[htb]
\centering
\caption{Manual assessment of code-mix detection tools precision in Hindi and Slovene.}
\label{tab:code-mixing-analysis}
\begin{tabular}{lcc}
\toprule
Dataset & Precision & Tool\\ 
\midrule
IIITH & $0.860$ & CodeSwitch\\
Hinglish Hate & $0.900$ & CodeSwitch\\ 
Sentiment15\textsubscript{SL} & $0.326$ & langdetect \\ 
IMSyPP\textsubscript{SL} & $0.201$ & langdetect \\ 
\bottomrule 
\end{tabular}
\end{table}

\Cref{tab:code-mixing-analysis} shows the manual evaluation of the precision obtained by our code-mix detection tools. For Hindi and Slovene datasets, we manually checked 50 random samples flagged as code-mixed and 50 random samples flagged as not code-mixed by our tools. While the CodeSwitch library, which is based on contextual LLM, shows a promising performance, the langdetect library often attributes words to an arbitrary language where they exist without considering their contextual usage. This limitation becomes evident when the same word appears in multiple languages within a sentence.
For instance, consider the word "brat", which refers to a male sibling in both Croatian and Slovene. The langdetect tool, using the class-conditional independence assumption of naïve Bayes, struggles to accurately determine the appropriate language for "brat" based on its surrounding context.


Acknowledging these challenges and the need for better code-switching detection accuracy, we manually annotated 100 samples from each dataset, excluding Hindi and Russian languages. This exclusion was based on the CodeSwitch library's promising precision rate in Hindi and the lack of collaborators to manually annotate Russian datasets. The results of this manual annotation and code-mixing percentage are shown in the rightmost column of \Cref{tab:dataset-statistics}. The code-mixing percentages in \Cref{tab:dataset-statistics} for the two Russian datasets are calculated on the whole dataset using the langdetect library.

\section{Pre-trained large language models}\label{sec:Pre}

In this section, we describe the pre-trained LLMs which we use in our study. In \Cref{sec:new-lms}, we first describe four newly created LLMs trained on considerable amounts of non-standard language aimed at better processing Slovene-English and Hindi-English code-mixed language. Then, we describe other (existing) LLMs used in our evaluation in \Cref{sec:existing-lms}. All models are listed in \Cref{tab:model-summary}, where the newly introduced models are marked with an asterisk (*).

\begin{table}[htb]
\caption{Summary of used large language models. The models marked with * are newly introduced in this paper. Models marked with $\blacktriangle$ are declared as monolingual but have capabilities in multiple languages. M and B stand for millions and billions of parameters, respectively.}
\label{tab:model-summary}
\centering
\begin{tabular}{lccr}
\toprule
Model & Link & Languages & Parameters \\
\midrule
\multicolumn{4}{l}{\textbf{Massively multilingual models:}} \\
mBERTc-base & \href{https://hf.co/bert-base-multilingual-cased}{link} & 104  & 178M \\
XLM-R-base &  \href{https://hf.co/xlm-roberta-base}{link} & 100  & 278M \\
TwHIN-BERT-base & \href{https://hf.co/Twitter/twhin-bert-base}{link}  & 100 & 279M \\ \hline

\multicolumn{4}{l}{\textbf{Few-lingual models:}} \\
SlEng-BERT* & \href{https://hf.co/cjvt/sleng-bert}{link} & sl, en & 117M \\
SloBERTa-SlEng* & \href{https://hf.co/cjvt/sloberta-sleng}{link} & sl, en & 117M \\
MuRIL-en-hi-codemixed* & \href{https://hf.co/cjvt/muril-en-hi-codemixed}{link} & hi, en & 117M \\
RoBERTa-en-hi-codemixed* & \href{https://hf.co/cjvt/roberta-en-hi-codemixed}{link} & hi, en & 117M \\
MuRILc-base & \href{https://hf.co/google/muril-base-cased}{link} & en + 16 in  & 238M  \\


HingRoBERTa & \href{https://hf.co/l3cube-pune/hing-roberta}{link} & hi, en & 278M \\
CroSloEngual BERT & \href{https://hf.co/EMBEDDIA/crosloengual-bert}{link}  & sl, hr, en & 124M \\ \hline

\multicolumn{4}{l}{\textbf{Monolingual models:}} \\
SloBERTa & \href{https://hf.co/EMBEDDIA/sloberta}{link} & sl & 111M \\
CamemBERT-base & \href{https://hf.co/camembert-base}{link} & fr & 111M \\



TamilBERT & \href{https://hf.co/l3cube-pune/tamil-bert}{link} & ta & 238M \\

RuBERTc-base & \href{https://hf.co/DeepPavlov/rubert-base-cased}{link} & ru & 178M \\ \hline

\multicolumn{4}{l}{\textbf{Monolingual generative models:}} \\
GPT3 $\blacktriangle$ & \href{https://openai.com/blog/gpt-3-apps}{link} & en & 175B \\

Llama2-7B $\blacktriangle$ & \href{https://hf.co/meta-llama/Llama-2-7b}{link} & en & 7B \\[-1pt]
\bottomrule
\end{tabular}
\end{table}

\subsection{New code-mixed language models}
\label{sec:new-lms}

We trained four new large language models on considerable amounts of non-standard language with the intention of using them for affective computing tasks and code-mixed processing. All the new models are masked LLMs utilizing transformer architecture \cite{transformer2017}, with about 117 million parameters each. The models are bilingual, two being Hindi-English and two Slovene-English. 
Two models were trained from scratch, and two were further trained from existing models on new data. Their names and classification are shown in \Cref{tab:cmModels}.

\begin{table}[htb]
\centering
\caption{The names and properties of four newly trained bilingual masked language models.}
\label{tab:cmModels}
\begin{tabular}{ccc}
\toprule
Languages & From scratch  & From existing \\ 
\midrule
Hindi-English & RoBERTa-en-hi-codemixed & MuRIL-en-hi-codemixed \\
Slovene-English & SlEng-BERT & SloBERTa-SlEng \\ 
\bottomrule 
\end{tabular}
\end{table}

Each of the newly trained models has 12 transformer encoder layers, equal in architecture and roughly equal in number of parameters to the base-sized BERT \cite{bert} and RoBERTa \cite{RoBERTa} models. The models support a maximum sequence length of 512 tokens. The pre-training task was masked language modeling, with no other tasks (e.g., next-sentence prediction). The models are publicly available on the HuggingFace repository of the Centre for Language Resources and Technologies of the University of Ljubljana\footnote{\href{https://huggingface.co/cjvt}{https://huggingface.co/cjvt}}. Their short description is provided below.

\subsubsection{Slovene-English}
The two new Slovene-English models \emph{SlEng-BERT} and \emph{SloBERTa-SlEng} are trained on Slovene and English conversational, non-standard, and slang language corpora. Concretely, they are trained on English and Slovene tweets \cite{janes-tag}, Slovene part of the web crawl corpus MaCoCu \cite{macocu-dataset}, the corpus of moderated web content Frenk \cite{frenk-datasets}, and a small subset of the English OSCAR corpus \cite{oscar-dataset}. The size of the English and Slovene corpora used is approximately equal. In total, the training data contains about 2.7 billion words, which were tokenized into 4.1 billion subword tokens prior to training. Both models share the same vocabulary and input embeddings, containing \num{40000} subword tokens.

Using the dataset, the models were trained using two different training regimes: SlEng-BERT was trained from scratch for $40$ epochs while SloBERTa-SlEng was initialized using the SloBERTa \cite{sloberta} Slovene monolingual masked LLM and further pre-trained for two epochs.

\subsubsection{Hindi-English}
The two new Hindi-English models \emph{RoBERTa-en-hi-codemixed} and \emph{MuRIL-en-hi-codemixed} are trained on English, Hindi, and code-mixed English-Hindi corpora. The corpora used consist of primarily web-crawled data, including code-mixed tweets, focusing on conversational language and the COVID-19 pandemic. The training corpora contain about 2.6 billion words, which were tokenized into 3.4 billion subword tokens prior to training.
Both models share the same vocabulary and input embeddings, containing \num{40000} subword tokens.

Similarly as for the Slovene-English models, the models were trained using two different training regimes.
The RoBERTa-en-hi-codemixed model was trained from scratch for 40 epochs while MuRIL-en-hi-codemixed was initialized using pre-trained MuRIL multilingual masked LLM \cite{khanuja2021muril} and further pre-trained for two epochs. 

\subsection{Existing language models}
\label{sec:existing-lms}

In addition to four newly introduced bilingual LLMs, we evaluate several existing massively multilingual, few-lingual, and monolingual LLMs which we describe next. Our newly introduced models are variants of the encoder-only BERT and RoBERTa LLMs, and so are most of the other models, but we also include two massive decoder-only LLMs due to their strong general performance. In addition to the general domain models, we test multiple tweet domain-adapted LLMs as they are specialized for handling social media texts that commonly include code-mixed language. Below, we split their descriptions into three groups: massively multilingual models, few-lingual models, monolingual models, and generative monolingual models.

\subsubsection{Massively multilingual models}
\label{sec:massively-multilingual-lms}

We consider three massively multilingual LLMs trained on 100 or more languages.

\textbf{Multilingual BERT} (mBERT) \cite{bert} is a multilingual masked LLM based on the BERT architecture \cite{bert}. It was trained on Wikipedia dumps in 104 languages with the largest Wikipedia size. We use the cased base-size version of this model and refer to it as mBERTc-base.

\textbf{XLM-RoBERTa} \cite{xlm-roberta} is a multilingual masked LLM based on the RoBERTa architecture \cite{RoBERTa}, trained on Wikipedia dumps and web crawl data in 100 languages. We use the base-size version and refer to the model as XLM-R-base.

\textbf{TwHIN-BERT} \cite{twhin-bert} is a multilingual masked LLM based on the BERT architecture \cite{bert}. It was trained on tweets in 100 languages. In addition to masked language modeling, it was trained using a contrastive social loss, the goal of which is to learn if two tweets appeal to similar users. We use the base-size version and refer to the model as TwHIN-BERT-base. 

\subsubsection{Few-lingual models}

As massively multilingual models cover a wide range of languages, their vocabulary and tokenization are not adapted to any specific language, which makes them inferior to LLMs incorporating fewer languages for many tasks \cite{ulcar2021training}. We consider three few-lingual LLMs covering considerably fewer languages than massively multilingual LLMs.

\textbf{MuRIL} \cite{khanuja2021muril} is a multilingual masked LLM based on the BERT architecture \cite{bert}. It was trained on large Indian corpora consisting of English and 16 Indian languages, including Hindi and Tamil which we consider in this work. We use the cased base-size version of this model and refer to it as MuRILc-base.


\textbf{HingRoBERTa} \cite{nayak-joshi-2022-l3cube} is a bilingual masked LLM based on RoBERTa architecture \cite{RoBERTa}. It was fine-tuned on a large corpus of Hindi-English tweets. HingRoBERTa has demonstrated competitive performance across various downstream tasks compared to other models trained on Hindi-English code-mixed datasets, as evidenced in research by Nayak et al. \cite{nayak2022l3cube}

\textbf{CroSloEngual BERT} \cite{csebert} is a trilingual masked LLM based on the BERT architecture \cite{bert}.
It was trained on a mixture of news articles, and web-crawled data in Croatian, Slovene, and English.

\subsubsection{Monolingual models}
While we hypothesize that multilingual LLMs might be preferable for code-mixed affective tasks, we also test monolingual models. They might be competitive and familiar with languages other than their main language, as a small number of other languages is likely to be present in all monolingual training corpora due to their huge size and likely presence in the news, textbooks, and social media.  We consider four monolingual masked LLMs covering specific languages.

\textbf{CamemBERT}\cite{martin2019camembert} is a French monolingual masked LLM based on the BERT architecture \cite{bert}. It was trained using a whole-word masking version of the masked language modeling objective on web-crawled data. We use the base-size version and refer to the model as CamemBERT\textsubscript{base}.

\textbf{SloBERTa} \cite{sloberta} is a Slovene monolingual masked LLM based on the CamemBERT architecture \cite{martin2019camembert}. It was trained on a union of five Slovene corpora containing news articles, web-crawled data, tweets, academic language, and parliamentary data.



\textbf{TamilBERT} \cite{joshi2023l3cubehindbert} is a Tamil monolingual masked LLM based on the BERT architecture \cite{bert}. It was trained on a Tamil monolingual corpus. 


\textbf{RuBERT} \cite{rubert} is a Russian monolingual masked LLM based on the BERT architecture \cite{bert}. It was trained on the Russian Wikipedia dump and news articles. We use the cased base-size version and refer to the model as RuBERTc\textsubscript{base}.


\subsubsection{Monolingual generative models}
 While all other models in our evaluation are masked LLMs using only the encoder stack of the transformer architecture, for comparison, we also include two popular and considerably larger generative LLMs that use only the decoder stack of the transformer architecture. The GPT3 and Llama2 models we describe below are primarily trained on English, although they have seen and are capable of processing other languages.
 
\textbf{GPT-3} \cite{brown2020language} is a language model trained on a vast amount of text and code. This vast training dataset allows it to perform a variety of tasks, including generating different creative text formats, translating languages, and answering questions in an informative way. The specific inner workings of this model remain undisclosed.

\textbf{Llama2} \cite{touvron2023llama} is an English monolingual autoregressive LLM. The details about its training data are scarce: the authors state that it was trained on mostly English data from publicly available sources. We use the version with 7B parameters and refer to the model as Llama2-7B.


\section{Evaluation}
\label{sec:eval}

In this section, we evaluate the described language models on multiple code-mixed datasets. 
We first describe the experimental setup in \Cref{sec:experimental-setuo} and continue with the analysis of results in \Cref{sec:results}. In \Cref{sec:results-codemixed-only} we further analyze the results, specifically focusing on the effect of code-mixed language. 

\subsection{Experimental setup}
\label{sec:experimental-setuo}

In our model evaluation process, we split the data into training, validation, and testing sets randomly in the proportion 60\%:20\%:20\%, using stratification across the class labels. However, for the GPT 3 and Lamma2 models, we implemented a nuanced approach to manage costs and streamline performance. For GPT, we selected a random subset of 500 samples from the training dataset and 100 from the testing dataset. Similarly, for Lamma2, we optimized output generation time by reducing the testing dataset to 300 samples.
Subsequently, we reported every model's macro $F_1$ score on the test set as a comprehensive metric for overall effectiveness.
To enhance the quality of the input data, we conducted preprocessing steps that involved the removal of special characters, such as '@', and trailing whitespaces from the text. 
To assure the reproducibility of our results, we share our code online\footnote{\href{https://github.com/matejklemen/sentiment-hate-speech-with-code-mixed-models}{https://github.com/matejklemen/sentiment-hate-speech-with-code-mixed-models}}. 
We provide descriptions of the finetuning process for different model groups next: in \Cref{sec:fine-tuning-BERT} for BERT-like models, in \Cref{sec:fine-tuning-GPT} for the GPT3 model, and in \Cref{sec:fine-tuning-Llama} for the Llama2 model.


\subsubsection{Fine-tuning BERT-like models}
\label{sec:fine-tuning-BERT}

We evaluate the models in a discriminative (classification) setting, meaning we fine-tune the models to discriminate between the two (in offensive language identification) or three (in sentiment analysis) unique classes.
To enable batch processing, we truncate and pad all input sequences to a maximum length of 512 subword tokens.
We optimize the models using the AdamW \cite{adamw} optimizer with the learning rate $5 \cdot 10^{-5}$ for up to five epochs, maximizing the batch size based on the available GPU memory. On Slovene datasets, we fine-tune the models for up to ten epochs as we noticed some models did not converge after five epochs.
Our fine-tuning settings are selected as reasonable defaults instead of using a thorough hyperparameter optimization, and largely follow existing practices for fine-tuning BERT-like models \cite{tran2019bert} \cite{McCormick2019BERT}. 

\subsubsection{Fine-tuning GPT3 Model}
\label{sec:fine-tuning-GPT}

\begin{figure}[h]
    \centering
    \includegraphics[width=1\columnwidth]{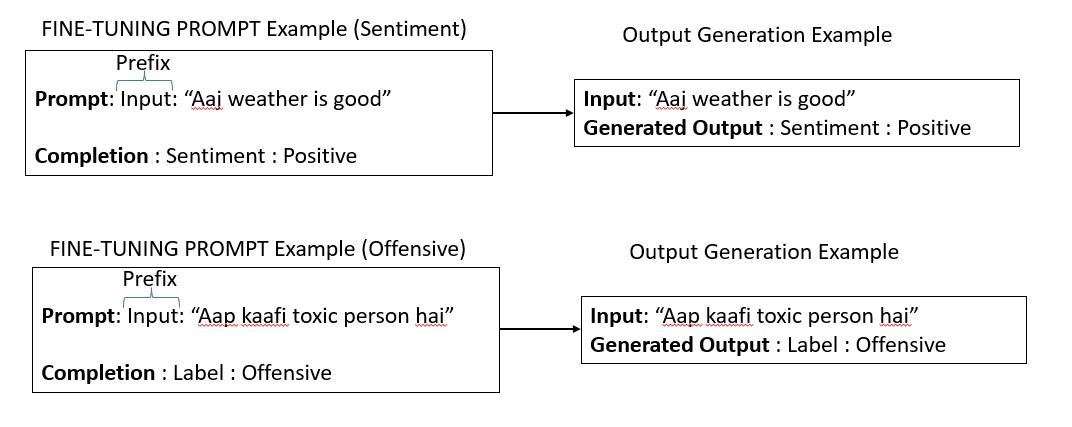}
    \caption{Example prompts and output templates for sentiment anaysis and offensive language detection tasks.}
    \label{fig:prompt-example}
\end{figure}

The evaluation for generative models necessarily differs from BERT-like classification models. Specifically, we fine-tune the GPT3 model to generate textual classes. In this process, GPT-3 is trained to map input text to output text using our prompt structure. 
We fine-tune the model for 800 prompt completion pairs for  2 epochs using the OpenAI API on randomly sampled training subsets. During the generation phase, we set the temperature to $0.1$ and the learning rate to $0.1$
During fine-tuning, each example is transformed into a prompt
 by appending the text to the prefix "Input:". The model is optimized to produce text with the prefix "Sentiment is:" (or "Label is:" in offensive language identification), followed by the predicted class. The generated output takes the form of 'Sentiment: {class}' or 'Label: {class}'.
%
 An example of a prompt and the output template are shown in Figure \ref{fig:prompt-example}, for a code-mixed Hindi-English input.

As shown, we fine-tune the GPT3 model to generate textual classes. Initially, we fine-tuned the Curie variant of GPT3 using our prompt structure and hyperparameters as previously described. However, due to Curie's deprecation, we transitioned to the davinci-002 variant for completing our tests on the Hindi datasets. For fine-tuning the davinci-002 model, we employed a temperature setting of $0.1$ and logit\_bias. 

\subsubsection{Fine-tuning the Llama2 models}
\label{sec:fine-tuning-Llama}
Llama2 generative model was fine-tuned with parameter-efficient fine-tuning method QLoRA\cite{dettmers2023qlora}, utilizing 16-bit precision quantization with a learning rate of $2 \cdot 10^{-4}$. The idea of parameter-efficient fine-tuning techniques \cite{lester2021power} is to selectively fine-tune a limited set of additional model parameters, significantly reducing both computational and storage expenses associated with the fine-tuning process.


The same as the GPT3 model, we evaluate the Llama-2 model in a generative setting, using prompts akin to those used in the GPT3 model (see \Cref{sec:fine-tuning-GPT}).

\subsection{Results on the full datasets}
\label{sec:results}

In this section, we present the results of different models on affective code-mixed tasks, sentiment prediction, and offensive language detection.  \Cref{tab:results-main} presents the results split by language, i.e. each subtable corresponds to a specific code-mixed datasets: Slovene (\Cref{tab:results-slovene}), Tamil (\Cref{tab:results-tamil}), Hindi (\Cref{tab:results-hindi}), French (\Cref{tab:results-french}), and Russian (\Cref{tab:results-russian}).

\begin{table}
\caption{ Macro $F_1$ scores on sentiment prediction and offensive language detection tasks achieved with massively multilingual, few-lingual, monolingual, and generative models across five primary languages. The best $F_1$ score for each language is displayed in \textbf{bold}.}

\label{tab:results-main}

\begin{subtable}{\columnwidth}
    \centering
    \caption{Results on Slovene datasets. The * marks the evaluation on a sample of 1000 examples.}
    \label{tab:results-slovene}
    \begin{tabular}{lcc}
    \toprule
    Model & 
    Sentiment15\textsubscript{SL} & 
    IMSyPP\textsubscript{SL} \\
    \midrule
    mBERTc-base & $0.617$ & $0.785$ \\
    XLM-R-base & $0.645$ & $0.811$ \\
    TwHIN-BERT-base & $0.619$ & $0.808$ \\ \hline

    CroSloEngual BERT & $0.660$ & $0.834$ \\
    SlEng-BERT & $\textbf{0.666}$ & $\textbf{0.850}$ \\
    SloBERTa-SlEng & $0.659$ & $0.845$ \\ \hline

    SloBERTa & $0.650$ & $0.840$ \\ \hline

    GPT3* & $0.597$ & $0.825$ \\
    Llama2-7B & $0.596$ & $0.821$ \\

    \bottomrule
    \end{tabular}
\end{subtable}

\hfill
\vspace{10pt}
\begin{subtable}{\columnwidth}
    \centering
    \caption{Results on Tamil datasets.}

    \label{tab:results-tamil}
    \begin{tabular}{lcccc}
    \toprule
    Model & \makecell{TamilCMSenti} & \makecell{TamilCMOLID.} 
    \\ 
    \midrule
    
    mBERTc-base & $0.749$ & $0.832$ \\ 
    XLM-R-base & $0.715$ & $0.778$ \\ 
    TwHIN-BERT-base & $\textbf{0.795}$ & $\textbf{0.896}$ \\ \hline 

    MuRILc-base & $0.688$ & $0.843$ \\ \hline 

    TamilBERT & $0.539$ & $0.772$ \\ \hline 

    GPT3 & $0.679$ & $0.781$ \\ 
    Llama2-7B & $0.620$ & $0.677$ \\ 
        

    \bottomrule
    \end{tabular}
\end{subtable}

\hfill
\vspace{10pt}
\begin{subtable}{\columnwidth}
    \centering
    \caption{Results on Hindi datasets. }
    \label{tab:results-hindi}
    \begin{tabular}{lcc}
    \toprule
    Model & \makecell{IIITH} 
    & \makecell{Hinglish Hate}\\
    \midrule
    
    mBERTc-base & $0.749$ & $0.711$ \\
    XLM-R-base & $0.739$ & $0.677$ \\
    TwHIN-BERT-base & $0.830$ & $0.733$ \\ \hline
    MuRILc-base & $0.698$ & $0.748$\\ \hline
    
    HingRoBERTa & $\textbf{0.854}$ & $\textbf{0.833}$ \\

    RoBERTa-en-hi-codemixed  & $0.792$ & $ 0.729$ \\
    MuRIL-en-hi-codemixed & $0.718$ & $0.709$ \\ \hline
    
    GPT3 & $0.660$ & $0.551$ \\
    Llama2-7B & $0.666$ & $0.595$ \\

    \bottomrule
    \end{tabular}
\end{subtable}

\hfill
\vspace{10pt}
\begin{subtable}{\columnwidth}
    \centering
    
    \caption{Results on French datasets. 
    }
    \label{tab:results-french}
    \begin{tabular}{lcccc}
    \toprule
    Model & FrenchBookReviews & FrenchOLID  \\ 
    \midrule
  
    mBERTc-base & $0.749$ & $0.915$ \\ 
    XLM-R-base & $0.644$ & $0.902$ \\ 
    TwHIN-BERT-base & $\textbf{0.759}$ & $\textbf{0.936}$ \\ \hline 
    CamemBERT & $0.696$ & $0.914$ \\ \hline 
    GPT3 & $0.743$ & $0.663$ \\ 
    Llama2-7B & $0.571$ & $0.714$ \\ 
    
    \bottomrule
    \end{tabular}
\end{subtable}

\hfill
\vspace{10pt}
\begin{subtable}{\columnwidth}
    \centering
    \caption{Results on Russian datasets.}
    \label{tab:results-russian}
    \begin{tabular}{lcccc}
    \toprule
    Model & {Sentiment15\textsubscript{RU}} & \makecell{RussianOLID} \\ 
    \midrule
    mBERTc-base & $\textbf{0.871}$ & $0.915$ \\ 
    XLM-R-base & $0.846$ & $0.902$  \\ 
    TwHIN-BERT-base & $0.851$ & $0.969$ \\ \hline 
    RuBERTc-base & $0.856$  & $\textbf{0.971}$ \\ \hline 
    GPT3 & $0.734$ & $0.665$ \\ 
    Llama2-7B & $0.677$ & $0.864$ \\ 

    \bottomrule
    \end{tabular}
   
\end{subtable}

\end{table}

On Slovene datasets the best results are achieved by the newly introduced SlEng-BERT model, achieving $F_1$ score $0.666$ on Sentiment15\textsubscript{SL} and $0.850$ on IMSyPP\textsubscript{SL}. In general, the results achieved by the Slovene monolingual (SloBERTa) and few-lingual (SlEng-BERT, SloBERTa-SlEng, CroSloEngual BERT) models are better than those achieved by the massively multilingual models and the significantly larger generative English models (GPT3 and Llama2). GPT3 and Llama2 achieve comparable scores on both datasets ($F_1$ scores $0.597$ and $0.596$ on Sentiment15\textsubscript{SL}, and $0.825$ and $0.821$ on IMSyPP\textsubscript{SL}), performing worse than massively multilingual models on Sentiment15\textsubscript{SL} and better than them on IMSyPP\textsubscript{SL}.

On Tamil datasets, TwHIN-BERT-base emerges as the top-performing model, achieving the $F_1$ score of 0.795 on the Tamil sentiment dataset and 0.896 on the offensive language dataset. Following closely is  mBERTc-base  with an $F_1$ score of 0.749 on the Tamil sentiment dataset and 0.832 on the offensive language dataset. XLM-R-base shows a competitive performance to mBERTc-base in sentiment analysis with an F1 score of 0.715, but falls short in hate speech detection tasks with an F1 score of 0.778. However, it should be noted that XLM-R-base has shown a trend of lower scores in Tamil in hate speech detection tasks in the research by Hossain et al.\cite{hossain2021nlp} where it shows lower results in Tamil when compared to other languages like English and Malayalam. The results indicate that specialized models trained for specific languages or tasks, such as TwHIN-BERT-base and  mBERTc-base , tend to outperform more generalized and larger models like GPT3 and Llama2-7B. GPT3, Llama2-7B, and TamilBERT show competitive but slightly lower performance across both datasets. 
MuRILc-base exhibit mixed performance across both the datasets. Their error analysis suggests that XLM-R-base struggles with distinguishing between HS (hate speech) and NHS (not hate speech) classes due to common code-mixed words, potentially affecting its performance. Additionally, the high class imbalance and biasness towards the not offensive label in both the cases (\cref{tab:label-distribution-olid}) may cause misclassification of offensive as not offensive.

On Hindi sentiment and offensive language tasks, HingRoBERTa emerges as the top model with $F_1$ scores of 0.854 and 0.833, showcasing the prowess of bilingual models. TwHIN-BERT-base and RoBERTa-en-hi-codemixed follow closely. 
The massively multilingual models mBERTc-base , MuRILc-base and XLM-R-base show comparable results. 
The second newly introduced model MuRIL-en-hi-codemixed performs worse than these models but better than generative models GPT3 and LlaMa2-7B. 

In French sentiment detection, massively multilingual TwHIN-BERT-base leads the race in sentiment analysis task with a $F_1$ score of 0.759 followed by mBERTc-base, GPT3, and monolingual CamemBERT. The XLM-R-base and Llama2-7B are trailing with considerable gaps. In the offensive speech detection task, several models have excelled in performance with TwHIN-BERT-base in the lead with a score of $F_1$ score of 0.936, followed by mBERTc-base, CamemBERT and XLM-R-base. 
GPT3 performance is considerably lower compared to other models.  

In Russian tasks, the monolingual model RuBERTc\textsubscript{base} leads in offensive speech detection with $F_1$ score of 0.971. Multilingual models mBERTc-base, XLM-R-base, and TwHIN-BERT-base perform competitively in sentiment analysis with mBERTc-base leading with $F_1$ score of 0.871. Massive generative models, GPT3 and Llama2-7B, lag behind.





Trying to draw more general conclusions, we observe two key findings. The best results are achieved by either bilingual models (Slovene and Hindi) or a model specialized for social media content (TwHIN-BERT-base on Tamil and French); the Russian language, without a bilingual model and with very low proportion of code-mixed text, is an exception here, but the model specialized for social media ((TwHIN-BERT-base) is very competitive. 

The advantage of bilingual models is two-fold. Firstly, their targeted focus on a specific language pair allows them to capture the intricacies of each language's vocabulary and grammar more effectively than massively multilingual models. This is reflected in the superior performance of HingRoBERTa in the Hindi-English datasets and SlEng-BERT in the Slovene code-mixed datasets compared to multilingual models, which may struggle with the nuances of code-mixing present in such data. 
Secondly, bilingual models typically have lower memory requirements compared to their multilingual counterparts. This translates to greater practical efficiency, making them more suitable for deployment in real-world applications, especially when dealing with resource constraints. 

\begin{figure}[h]
    \centering
    \begin{subfigure}{0.49\textwidth}
        \centering
        \includegraphics[width=\linewidth]{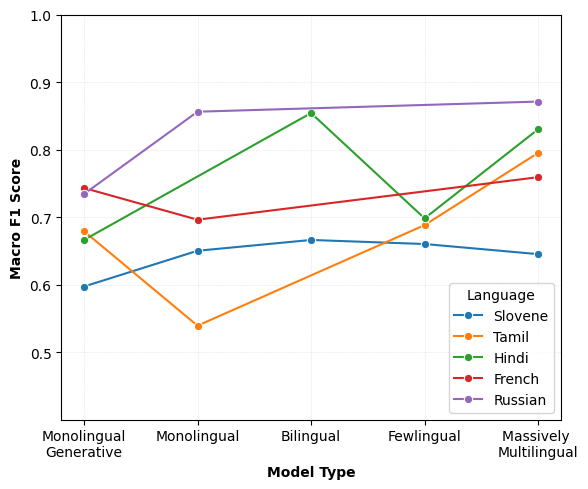}
        \caption*{a) Sentiment analysis.}
    \end{subfigure}
    \hfill
    \begin{subfigure}{0.49\textwidth}
        \centering
        \includegraphics[width=\linewidth]{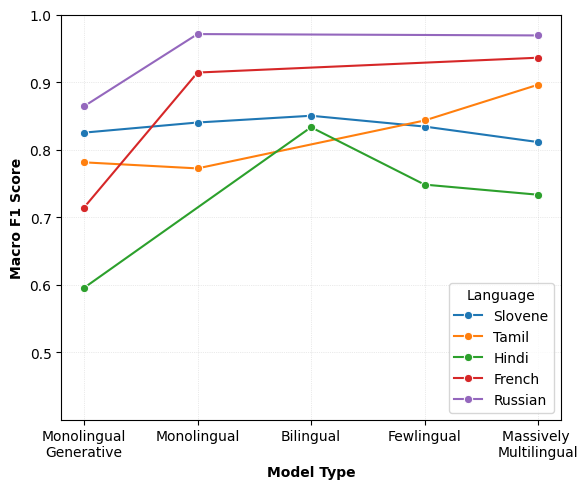}
        \caption*{b) Hate speech detection.}
    \end{subfigure}
    \caption{Comparing the performance of the best model in each model category for the used tasks.}
    \label{fig:Model-comparision}
\end{figure}

Figure \ref{fig:Model-comparision} presents a comparative analysis of various categories of models -— generative, monolingual, bilingual, few-lingual, and massively multilingual -— in the context of the two affective computing tasks. For this comparison, we considered the model with the best performance within each category. Figure \ref{fig:Model-comparision}a illustrates the results for sentiment analysis. Generative models, while performing adequately, lag behind other models. Monolingual models perform better but still lag behind other types of models. Bilingual models demonstrate superior performance where they exist, closely followed by massively multilingual models. Interestingly, a dip in performance is observed for few-lingual models. Figure \ref{fig:Model-comparision}b showcases the performance of the models in the offensive speech detection task. 
The trends show that generative models lag behind others, massively multilingual models either perform comparably to or fall below bilingual and fewlingual models, with bilingual models mostly maintaining a prominent position.

\vspace{\baselineskip}

\subsection{Results on the code-mixed subsets}
\label{sec:results-codemixed-only}
In this section, we analyze the performance of models separately for code-mixed and non-code-mixed subsets of their respective datasets. We select Slovene and Hindi as the representative languages and curate the datasets to isolate code-mixed examples.
For Hindi, we use the CodeSwitch library to separate code-mixed from non-code-mixed examples in the test sets.
For Slovene, where language detection tools perform poorly on code-switched text, we manually selected a subset of 1000 examples from the test set (identical to the sample used for GPT3 evaluation in \Cref{sec:results}) and manually annotated it for code-mixing. \Cref{sec:results-by-codemixing} provides a comparison of model performance on code-mixed (CM) and non-code-mixed (notCM) subsets.

On the code-mixed text, the overall best bilingual model SlEng-BERT, performs comparably or slightly worse to the best-performing monolingual SloBERTa model.
During our manual inspection of the code-mixing patterns, we find that code-mixing is very rarely the dominant cause for the target label, therefore, it is unsurprising that the code-mixed models do not perform significantly better than the general-purpose ones on the code-mixed subset. These findings might indicate a lesser impact of code-mixing in affective tasks as commonly assumed. 
However, to confirm such findings. a future more focused research of code-mixed text processing is necessary, using better datasets with carefully curated texts (e.g., in the form of contrast sets \cite{contrast-sets}).

Results on Hindi datasets show several intriguing patterns. The bilingual HingRoBERTa consistently demonstrates the best performance across all scenarios, showcasing its efficacy in capturing the nuances of both sentiment analysis and offensive speech detection in Hindi. The newly introduced models, RoBERTa-en-hi-codemixed and MuRIL-en-hi-codemixed, perform better in the code-mixed data subsets compared to the non-code-mixed subsets. This could be attributed to their training in Hinglish code-mixed corpora, giving them a certain advantage in handling code-mixed text. In contrast, GPT3 displays subpar results across both genres of tasks, indicating potential limitations in its adaptability to the complexities of Hindi text for sentiment and hate speech analysis. Interestingly, when considering offensive language detection, models mostly perform better in code-mixed scenarios. This phenomenon may be attributed to the possibility that multiple languages share similar hate speech tendencies, as suggested by Arango et al.\cite{arango2021cross}, who suggested the existence of common patterns in offensive speech across different languages.

\begin{table}
\caption{Separate results for code-mixed (CM) and not code-mixed data (notCM).}
\label{sec:results-by-codemixing}

\begin{subtable}{\columnwidth}
    \centering
    \caption{Separate results on Slovene datasets. }
    \label{tab:codemixing-perf-sl-en}
    \begin{tabular}{lcccc}
    \toprule
    Model & 
    \multicolumn{2}{c}{\textbf{Sentiment15\textsubscript{SL}}} & 
    \multicolumn{2}{c}{\textbf{IMSyPP\textsubscript{SL}}} \\
    & \textit{CM} & \textit{notCM} & \textit{CM} & \textit{notCM} \\
    & $N = 389$ & $N = 611$ & $N = 254$ & $N = 746$ \\
    \midrule
    mBERTc-base & $0.609$ & $0.623$ & $0.825$ & $0.763$ \\
    XLM-R-base & $0.631$ & $0.648$ & $0.828$ & $0.821$ \\
    TwHIN-BERT-base & $0.589$ & $0.633$ & $0.811$ & $0.822$ \\ \hline
    CroSloEngual BERT & $0.653$ & $0.671$ & $0.848$ & $0.853$ \\    
    SlEng-BERT & $0.659$ & $\textbf{0.690}$ & $0.874$ & $\textbf{0.865}$ \\ 
    SloBERTa-SlEng & $0.649$ & $0.661$ & $0.862$ & $0.852$ \\ \hline
    SloBERTa & $\textbf{0.666}$ & $0.651$ & $\textbf{0.876}$ & $0.854$ \\ \hline
    
    GPT3* & $0.575$ & $0.606$ & $0.831$ & $0.816$ \\
    Llama2-7B & $0.595$ & $0.598$ & $0.803$ & $0.840$ \\
    
    \bottomrule
    \end{tabular}
\end{subtable}

\hfill
\vspace{10pt}
\begin{subtable}{\columnwidth}
    \centering
    \caption{Separate results on Hindi datasets.
    }
    \label{tab:seq-cls-2-hi-en}
    \begin{tabular}{lcccc}
        \toprule
    Model & 
    \multicolumn{2}{c}{\textbf{IIITH}} & 
    \multicolumn{2}{c}{\textbf{Hinglish Hate}} \\
    & \textit{CM} & \textit{notCM} & \textit{CM} & \textit{notCM} \\
    & $N = 473$ & $N = 303$ & $N = 780$ & $N = 135$ \\
    \midrule
    mBERTc-base & $0.776$ & $ 0.779 $ & $0.725 $ & $0.645$ \\
    XLM-R-base & $0.822$ & $0.797$ & $0.680$ & $0.588$ \\
    TwHIN-BERT-base & $0.828$ & $0.832$ & $0.740$ & $0.685$ \\ \hline
    MuRILc-base & $0.707$ & $0.650$ & $0.673$ & $0.564$ \\ \hline
    HingRoBERTa & $\textbf{0.876}$ & $\textbf{0.867}$ & $\textbf{0.838}$ & $\textbf{0.831}$\\
    RoBERTa-en-hi-cm & $0.850$ & $0.787$ & $0.692$ & $ 0.564$\\
    MuRIL-en-hi-cm & $0.723$ & $0.669$ & $0.704$ & $0.625$\\ \hline
    GPT3 & 0.466 & 0.419  & 0.548 & 0.568\\
    Llama2-7B & $0.737$ & $0.688$ & $0.690$ & $0.720$ \\

    \bottomrule
    \end{tabular}
\end{subtable}

\end{table}

\section{Conclusion}\label{sec:conclusion}

Our research analyzed the performance of several types of large language models on two affective computing tasks in a code-mixed setting. A notable finding is the dominance of bilingual and multilingual models, specialized for social media texts, over general massively multilingual, few-lingual, and monolingual models across diverse language pairs. For instance, the bilingual HIndi-Engish HingRoBERTa model and the newly introduced Slovene-English  SlEng-BERT demonstrated the best $F_1$ scores in sentiment analysis and offensive speech detection for Hindi and Slovene, respectively.

A separate analysis of code-mixed and non-code-mixed data subsets showed slightly better performance of almost all models on code-mixed data compared to non-code-mixed data for both Slovene-English and Hindi-English code-mixing scenarios. While this might indicate that  certain affective role of code-mixing, more research is needed to confirm this hypothesis, especially as manual inspection showed relatively little impact of code-mixing in sentiment detection and offensive speech detection.

Our findings provide a foundation for future explorations. While our focus has predominantly been on sentiment analysis and offensive speech detection, future work shall extend beyond these realms to encompass a broader spectrum of affective tasks, including emotion and sarcasm detection. Diversifying the language pairs and refining fine-tuning strategies for low-resource settings are crucial steps forward.  The transition from research to practical applications involves testing sentiment analysis models and offensive speech detection models in real-world scenarios, further validating their utility.

\section{Acknowledgement}\label{sec:acknowledgement}
The work was partially supported by the Slovenian Research and Innovation Agency (ARIS) core research programme P6-0411, young researcher grant, as well as projects J7-3159, CRP V5-2297, L2-50070, BI-IN/22-24-015 and DST/ICD/lndo-Slovenia/2022/15(G).
We sincerely thank Alice Baudhuin for her expertise in language analysis, particularly in manually detecting code-mixing within our French datasets.

\bibliographystyle{plain}
\bibliography{references}  

\end{document}